# Accurate Optical Flow via Direct Cost Volume Processing

Jia Xu  René Ranftl  Vladlen Koltun

Intel Labs

## Abstract

*We present an optical flow estimation approach that operates on the full four-dimensional cost volume. This direct approach shares the structural benefits of leading stereo matching pipelines, which are known to yield high accuracy. To this day, such approaches have been considered impractical due to the size of the cost volume. We show that the full four-dimensional cost volume can be constructed in a fraction of a second due to its regularity. We then exploit this regularity further by adapting semi-global matching to the four-dimensional setting. This yields a pipeline that achieves significantly higher accuracy than state-of-the-art optical flow methods while being faster than most. Our approach outperforms all published general-purpose optical flow methods on both Sintel and KITTI 2015 benchmarks.*

## 1. Introduction

Optical flow estimation is a key building block of computer vision systems. Despite concerted progress, accurate optical flow estimation remains an open challenge due to large displacements, textureless regions, motion blur, and non-Lambertian effects. Tellingly, the accuracy of leading optical flow algorithms is behind the accuracy achieved for the related problem of stereo matching. This is despite the close structural similarity of the two problems: stereo matching can be viewed as a special case of optical flow.

The most successful methods for stereo matching and optical flow tend to follow different philosophies. Leading stereo methods treat the search space as a highly regular discrete structure and explicitly construct a complete representation of this structure, known as the cost volume [29, 39]. This enables the application of powerful global and semi-global optimization techniques that remove outliers and enforce coherence [16, 33]. In contrast, the cost volume for optical flow is four-dimensional and its explicit construction and processing have until recently been considered infeasible. For this reason, optical flow methods commonly rely on nearest neighbor search [25, 3, 12, 2] and coarse-to-fine analysis [28, 3].

Recent work has indicated that operating on the com-plete cost volume, à la stereo, is in fact feasible and that the regular structure of this volume supports the use of global optimization techniques [7]. However, the computational requirements of this approach appeared to render it impractical, due both to the construction of the cost volume and the optimization over it. It remained unclear whether we can translate the successful structure of state-of-the-art stereo processing pipelines to optical flow without incurring severe computational penalties.

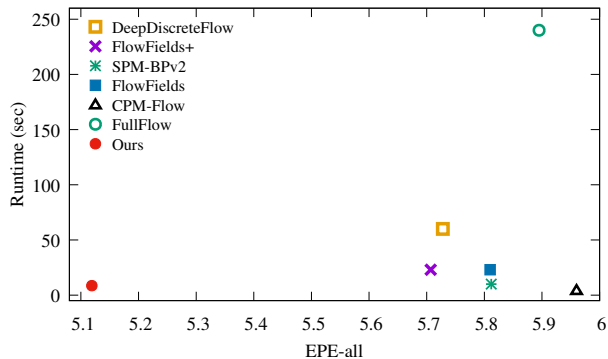

Figure 1. Accuracy versus running time on the Sintel benchmark. We compare to the top-ranking published optical flow methods. Our approach is significantly more accurate, while maintaining a competitive running time.

In this paper, we show that an optical flow algorithm can combine the convenience and accuracy of cost-volume processing with speed. Our work is based on learning an embedding into a compact feature space, such that matching scores between patches can be computed by inner products in this space. We show that the full four-dimensional cost volume can be constructed in a fraction of a second due to its regularity. We then exploit this regularity further by adapting semi-global matching [16] to the four-dimensional setting. Despite the size of the label space, its regularity exposes massive parallelism that can be harnessed to keep running times down. Additional postprocessing is performed by fitting homographies to image regions and using these to regularize the flow field.

The resulting pipeline achieves the highest reported accuracy on the Sintel benchmark [6] while maintaining a competitive running time. Our approach also significantly outperforms all published domain-agnostic optical flow methods on the KITTI 2015 benchmark [24], reducing the Fl-all error by 29.5% relative to the best prior work ("Patch-Batch" at the time of submission). Figure 1 illustrates the accuracy and running time of our approach in comparison to leading published methods. The presented approach even outperforms some recent methods that use additional domain-specific semantic supervision during training [31, 19], without using such additional supervision and at substantially lower running time.

## 2. Related Work

Optical flow estimation has made significant strides since its early days. In particular, the problem is largely solved when image motion is small [4, 32]. Recent work has therefore focused on the challenges brought up by large displacements [5]. Two specific challenges can be identified. First, patch appearance can change significantly with large motion. Second, large motions induce a correspondingly large potential search space that must be taken into account when correspondences are established. Recent advances in optical flow estimation can be categorized by how these challenges are addressed.

In estimating patch similarity, most approaches in the literature rely on hand-crafted matching functions and descriptors [5, 34, 25, 3, 7]. This had also been the case in stereo matching [17] until the recent popularization of matching functions based on convolutional networks [39, 22]. Such learned matching functions have recently been adopted for optical flow estimation [12, 13, 2]. Like these recent works, we use a learned matching function.

The second challenge is the large size of the search space. Many approaches use nearest neighbor search to restrict the domain of the algorithm to sparse matches, albeit at the cost of regularity [25, 3, 12, 13, 2]. Another approach that bypasses cost volume construction is multi-scale analysis of correspondence fields [3, 28]. We take a more direct tack and simply construct the cost volume. In this we are inspired by Full Flow, which demonstrated that the regularity of the four-dimensional cost volume affords significant benefits [7]. Our work demonstrates that cost volume processing is not antithetical to speed. Our approach achieves significantly higher accuracy and is more than an order of magnitude faster than Full Flow [7]. Surprisingly, it is also faster than almost all aforementioned methods that avoid cost volume construction, in addition to being significantly more accurate.

A recent line of work trains neural networks to directly estimate optical flow, stereo, and scene flow [9, 23]. The end-to-end nature of this approach is appealing, and the trained networks are very fast. However, the networks employed in these works have tens of millions of parameters, and consequently require a lot of external training data. In contrast, our network is very compact (112K parameters) and is two orders of magnitude smaller than both FlowNetS and FlowNetC [9]. As a result, we were able to train our network from scratch using only the training data in each benchmark (Sintel and KITTI 2015, respectively), without dataset augmentation. Such compact networks are advantageous in practical deployment [14]. Furthermore, our pipeline is highly modular, and different components (matching, cost volume processing, postprocessing) can be readily analyzed and upgraded.

Another recent family of methods takes advantage of domain-specific knowledge and combines optical flow with semantic segmentation [31, 19, 2]. This is particularly relevant in the automotive domain, where human-annotated ground-truth semantic label maps are available alongside compatible optical flow datasets. Such methods have yielded the highest accuracy to date on automotive datasets, but are limited in their generalization ability. As a symptom, these methods do not report results on the Sintel dataset. Our approach does not use semantic information and is agnostic to the domain. Nevertheless, it achieves higher accuracy than some of the domain-specific methods in the automotive domain, while retaining generality.

## 3. Overview

Following common recent practice, much of our pipeline operates on moderately downsampled images [7, 12, 13, 25]. Specifically, feature extraction, cost volume construction, and cost volume processing operate on images downsampled by a factor of three in each dimension. After cost volume processing, the correspondence field is upsampled to full resolution, then inpainted and refined.

**Cost volume construction.** Let $\mathbf{I}^1$ and $\mathbf{I}^2$ be two downsampled color images of resolution $M \times N$, represented as matrices in $\mathbb{R}^{MN \times 3}$. The images are additionally normalized to have zero mean and unit standard deviation. We begin by computing a $d$-dimensional feature vector for each pixel. Each image is processed by a convolutional network that produces feature vectors for all pixels jointly, yielding corresponding feature space embeddings $\mathbf{F}^1, \mathbf{F}^2 \in \mathbb{R}^{MN \times d}$. The four-dimensional cost volume is then populated by distances between pairs of feature vectors $(\mathbf{f}^1, \mathbf{f}^2)$, where $\mathbf{f}^1 \in \mathbf{F}^1$ and $\mathbf{f}^2 \in \mathbf{F}^2$. A simple property of the Euclidean metric allows constructing the cost volume in parallel using highly efficient vector products. This stage is described in Section 4.

**Cost volume processing.** The cost volume produced in the previous stage can be directly used to estimate optical flow via winner-take-all assignment, without any further processing. Our experiments will demonstrate that this already yields surprisingly good results. However, the cost volume can be processed to increase accuracy further by re-moving outliers and regularizing the estimated flow. To this end, we use an adaptation of semi-global matching (SGM) to four-dimensional cost volumes. This adaptation retains the regular and parallel operation of original SGM and can thus be executed efficiently. This is described in Section 5.

**Postprocessing.** We compute the forward flow from $\mathbf{I}^1$ to $\mathbf{I}^2$ and the backward flow from $\mathbf{I}^2$ to $\mathbf{I}^1$ and remove inconsistent matches. The remaining matches are lifted to the original resolution, resulting in a semi-dense correspondence field. We now use inpainting and variational refinement to obtain a dense subpixel-resolution flow field. To this end, we combine the EpicFlow interpolation scheme [27] with a complementary scheme that segments the images based on low-level edge cues and fits homographies to image segments. These homographies assist in inpainting large occluded regions. This is described in Section 6.

## 4. Feature Embedding

We learn a nonlinear feature embedding using a convolutional network [20]. Our goal is to embed image patches into a compact and discriminative feature space that is robust to geometric and radiometric distortions encountered in optical flow estimation. An additional requirement is that feature space embeddings as well as distances in this space can be computed extremely efficiently. This will allow rapid construction of the 4D cost volume. With these goals in mind, we design a small fully-convolutional network that embeds raw image patches into a compact Euclidean space.

**Parameterization.** Our network has 4 convolutional layers. Each of the first three layers uses 64 filters. Each convolution is followed by a pointwise truncation $\max(\cdot, 0)$ [26]. All filters are of size $3 \times 3$. We do not stride, pool, or pad. The last layer uses $d$ filters and their output is normalized to produce a unit-length feature vector $\mathbf{f} \in \mathbb{R}^d$ such that $\|\mathbf{f}\|_2 = 1$.

The network has a relatively small receptive field of $9 \times 9$ pixels, which has proven to be effective for stereo estimation [39]. Since this network operates on downsampled images, as described in Section 3, the induced receptive field in the original images is $27 \times 27$.

The dimensionality $d$ of the feature space poses a tradeoff between its expressive power and the computational cost of computing distances in this space. We will show in Section 7 that a surprisingly low dimensionality supports highly discriminative embeddings.

**Training.** We train a convolutional network $f : \mathbb{R}^{9 \times 9} \to \mathbb{R}^d$ that embeds input patches into the feature space. Let $\boldsymbol{\theta}$ be the parameters of the network. Let $\mathcal{D} = \{(\mathbf{x}_i^a, \mathbf{x}_i^p, \mathbf{x}_i^n)\}_i$ be a set of triplets of patches such that $\mathbf{x}_i^a$ is known to be more similar to $\mathbf{x}_i^p$ than to $\mathbf{x}_i^n$, for all $i$.

We optimize the embedding using the triplet loss [30, 36]:

$$\mathcal{L}(\boldsymbol{\theta}) = \frac{1}{|\mathcal{D}|} \sum_{i=1}^{|\mathcal{D}|} \big[ m + \|f(\mathbf{x}_i^a; \boldsymbol{\theta}) - f(\mathbf{x}_i^p; \boldsymbol{\theta})\|^2 \\ - \|f(\mathbf{x}_i^a; \boldsymbol{\theta}) - f(\mathbf{x}_i^n; \boldsymbol{\theta})\|^2 \big]_+. \quad (1)$$

To harvest the dataset $\mathcal{D}$ of training triplets, we use ground-truth optical flow, which is assumed to be provided for a training set of image pairs. For each image pair, we randomly sample an anchor $\mathbf{x}^a$ from the first image and use the ground-truth flow to obtain the corresponding positive patch $\mathbf{x}^p$ in the second image. To obtain corresponding negative examples $\mathbf{x}^n$, we randomly sample three patches in the second image at distances between 1 and 5 pixels from the center of $\mathbf{x}^p$. This yields three training triplets. This procedure can be repeated to produce hundreds of millions of training triplets from standard optical flow datasets.

Training is performed using SGD with momentum 0.9. For efficiency, the dataset $\mathcal{D}$ is constructed online during training, by a parallel thread that continuously samples new triplets and constructs mini-batches that are passed on to the solver. We use a batch size of 30K triplets to balance the execution of the data generation thread and the solver. 10K iterations are performed with a learning rate of $10^{-1}$, followed by 10K iterations with a learning rate of $10^{-2}$, followed by 20K iterations with a learning rate of $10^{-3}$. We do not use data augmentation or hard negative mining. The training set contains hard triplets by construction, since the positive and negative patches may be as little as one pixel apart.

**Cost volume construction.** For testing, we take the advantage of the fully-convolutional nature of the network and compute a feature embedding for all pixels in an image in a single forward pass through the network. Since the features are normalized to unit length, the matching cost can be computed using vector products, as shown below. This enables highly efficient cost volume construction.

Recall that our input images are $\mathbf{I}^1, \mathbf{I}^2 \in \mathbb{R}^{MN \times 3}$. Let $\mathbf{V} \in \mathbb{R}^{MN \times 2}$ be a flow field between $\mathbf{I}^1$ and $\mathbf{I}^2$. Let $\mathbf{V}_p$ be the flow at pixel $p \in [1, \ldots, MN]$. We assume that the search space is discrete and rectangular. Specifically, we assume that $\mathbf{V}_p \in \mathcal{R}^2$, where

$$\mathcal{R} = \{-r_{\max}, -r_{\max}+1, \ldots, 0, \ldots, r_{\max}-1, r_{\max}\}$$

and $r_{\max}$ is the maximal displacement. Let $\mathbf{F}^1, \mathbf{F}^2 \in \mathbb{R}^{MN \times d}$ denote the feature space embeddings of whole images $\mathbf{I}^1$ and $\mathbf{I}^2$, respectively. Let $\mathbf{C} \in \mathbb{R}^{MN \times |\mathcal{R}|^2}$ be the optical flow cost volume. Every entry in $\mathbf{C}$ can be computed as

$$\mathbf{C}(p, \mathbf{v}) = 1 - \left(\mathbf{F}_p^1\right)^\top \mathbf{F}_{p+\mathbf{v}}^2. \quad (2)$$

Here we take advantage of the connection between the Euclidean distance and the dot product. Since the feature vectors $\mathbf{F}_p^1$ and $\mathbf{F}_{p+\mathbf{v}}^2$ are normalized,

$$1 - \left(\mathbf{F}_p^1\right)^\top \mathbf{F}_{p+\mathbf{v}}^2 = \frac{1}{2} \left\|\mathbf{F}_p^1 - \mathbf{F}_{p+\mathbf{v}}^2\right\|^2. \quad (3)$$

This allows us to populate the cost volume using vector products, which can be evaluated in parallel.

It is easy to see that each entry in the cost volume can be computed in time $O(d)$ and the cost volume as a whole can be constructed in time $O(MN\mathcal{R}^2 d)$ (without taking parallelism into account). The dimensionality $d$ of the feature space thus has a direct effect on the computational cost of cost volume construction: reducing the dimensionality by an order of magnitude accelerates cost volume construction by an order of magnitude.

## 5. Cost Volume Processing

Recent work has shown that approximate global optimization over the full 4D cost volume can be performed using parallelized message passing and nested distance transforms [7]. However, the cost of this approach is still prohibitive: minutes per image after optimization [7]. We develop an alternative solution based on SGM, a technique that has been widely adopted in stereo processing [16]. SGM has become a common stand-in for more costly Markov random field optimization in stereo processing pipelines, due to its robustness and parallelism. For example, it is a core part of the successful recent pipeline of Žbontar and LeCun, which significantly advanced the state of the art in the area [39]. A strong connection between SGM and full Markov random field optimization is known, providing theoretical backing for what was originally a heuristic [10].

While restricted forms of SGM have been applied to optical flow before [15, 2], we are not aware of work that shows that SGM is tractable, efficient, and accurate when applied to the full four-dimensional cost volume. We now describe our adaptation of SGM, which we refer to as Flow-SGM. Let $\mathcal{N}(p)$ denote the set of spatial neighbors of pixel $p$. We adopt a simple 4-connected neighborhood structure. Define the discrete energy of the optical flow field $\mathbf{V}$ as

$$E(\mathbf{V}) = \sum_p \Bigg( \sum_{q \in \mathcal{N}(p)} P_1 [\|\mathbf{V}_p - \mathbf{V}_q\|_1 = 1]$$
$$+ \sum_{q \in \mathcal{N}(p)} P_2^{p,q} [\|\mathbf{V}_p - \mathbf{V}_q\|_1 > 1] + \mathbf{C}(p, \mathbf{V}_p) \Bigg), \quad (4)$$

where $[\cdot]$ denotes the Iverson bracket, and $P_1$ and $P_2^{p,q}$ are regularization parameters. We set $P_1$ to a fixed constant value and set

$$P_2^{p,q} = \begin{cases} P_2/Q & \text{if} \quad \|\mathbf{I}_p^1 - \mathbf{I}_q^1\| \geq T \\ P_2 & \text{else,} \end{cases} \quad (5)$$

where the threshold $T$ together with the constants $P_2$ and $Q$ are used to support edge-aware smoothing of the cost volume. Energy (4) is similar to the classical definition of the SGM objective [16]. The difference is that the displacement $\mathbf{V}_p$ is two-dimensional rather than scalar. In turn the definition of the regularization terms is based on two-dimensional neighborhoods, which is reflected in the $\ell_1$-norm based distance $\|\mathbf{V}_p - \mathbf{V}_q\|_1$. The similarity to the classical SGM objective is intentional since this type of energy can be processed efficiently using scanline optimization, even in the case of 2D displacements.

Flow-SGM approximately minimizes energy (4) by breaking the energy into independent paths, which can be globally minimized using dynamic programming. For each path, a cost $L_r(p, \mathbf{V}_p)$ is computed as

$$L_r(p, \mathbf{V}_p) = \mathbf{C}(p, \mathbf{V}_p) + S(p, \mathbf{V}_p) \\ - \min_i \left( L_r(p-r, i) + P_2^{p, p-r} \right), \quad (6)$$

where the contribution of the smoothness penalty $S(p, \mathbf{V}_p)$ is recursively computed as

$$S(p, \mathbf{V}_p) = \min \begin{cases} L_r(p-r, \mathbf{V}_p) \\ \min_{\hat{\mathbf{v}} \in \mathcal{N}(\mathbf{V}_p)} L_r(p-r, \hat{\mathbf{v}}) + P_1 \\ \min_i L_r(p-r, i) + P_2^{p, p-r}. \end{cases} \quad (7)$$

Here $r$ denotes the direction of traversal of the path. Note that in contrast to classical SGM, the computation of the penalty for switching by one discretization step is computed over a two-dimensional neighborhood. In practice, multiple path directions $r$ are used and the corresponding costs $L_r(p, \mathbf{V}_p)$ are accumulated into a filtered cost volume $L(p, \mathbf{V}_p)$. We use the four cardinal path directions: two horizontal and two vertical. The final optical flow estimate is given by picking the flow corresponding to the smallest cost in the filtered cost volume for each pixel. We compute the flow in both directions and use a consistency check to prune occluded or unreliable matches. The resulting high-quality matches are then passed on for postprocessing as described in the next section.

We implemented Flow-SGM on the GPU to make use of the massive amount of parallelism inherent in the algorithm. Because of the size of the cost volume, economical use of memory is important. To this end, we rescale and bin the values $\mathbf{C}(p, \mathbf{V}_p)$ to an 8-bit integer range. Since the maximal value of $L(p, \mathbf{V}_p)$ is bounded [16], we can store the filtered cost volume using 16 bits per entry.

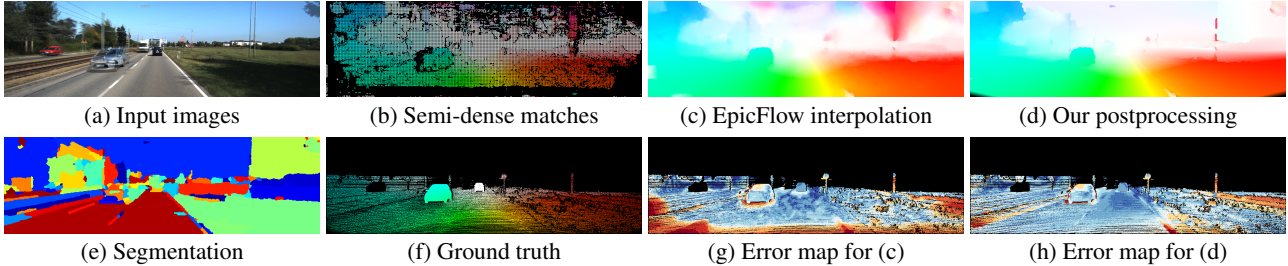

| (a) Input images | (b) Semi-dense matches | (c) EpicFlow interpolation | (d) Our postprocessing |
| (e) Segmentation | (f) Ground truth | (g) Error map for (c) | (h) Error map for (d) |

Figure 2. Postprocessing. (a) Superimposed input images. (b) Semi-dense matches provided as input to the postprocessing stage. (c) Dense and subpixel-resolution flow field produced by the EpicFlow interpolation scheme. (d) A flow field produced by our postprocessing stage, which incorporates homography-based inpainting. (e) Low-level segmentation used by our scheme. (f) Ground-truth flow between the input images. (g,h) Error maps corresponding to (c) and (d).

## 6. Postprocessing

Our starting point for converting semi-dense correspondences into a fully dense flow field is the EpicFlow interpolation scheme [27], which is commonly used for this purpose [3, 7, 12, 13, 25]. EpicFlow uses locally-weighted affine models to synthesize a dense flow field from semi-dense matches. We found that this scheme yields accurate results in areas where input matches are fairly dense, but is less reliable when large occluded regions must be filled. To address this, we develop a complementary interpolation scheme that greatly enhances inpainting performance in these regions.

We make use of the fact that large segments of optical flow fields can be characterized by planar homographies. This parameterization has been successfully applied in the context of scene flow, motion stereo, and optical flow [35, 37, 38]. The main challenge lies in identifying the extent of planar regions and making estimates robust and spatially consistent. Our key observation is that given high-quality semi-dense matches, it is relatively easy to identify these regions using the matches along with appearance information.

Our approach is based on a segmentation hierarchy combined with a greedy bottom-up fitting strategy. We compute an ultrametric contour map (UCM) [1] using a fast boundary detector [8]. A key property of UCM is that thresholding the map at different levels induces a segmentation hierarchy. We create a two-level hierarchy by thresholding the UCM at levels $t_1$ and $t_2$, where $t_2 > t_1$. We then fit homographies to the semi-dense matches belonging to segments in the finer level of the hierarchy using RANSAC [11]. We consider the homography a valid explanation for the flow in the segment if its inlier set is sufficiently large. We further aggregate larger segments by considering segments at the coarse level to be candidates for homography inpainting if the amount of inliers in their children was sufficiently large. For each such higher-level segment, we again robustly fit a homography and consider it valid if enough inliers are found.

For each segment with a valid homography, we use this homography to extrapolate the optical flow within the segment. All other segments are inpainted using the EpicFlow scheme.

Note that no semantic information is used. We rely on the same low-level edge cues as EpicFlow interpolation. As a consequence, our complementary inpainting scheme is just as broadly applicable. It adds little extra computation time but can greatly enhance the synthesized flow field in the presence of large occluded regions. This is illustrated in Figure 2 and will be evaluated in controlled experiments in Section 7.

## 7. Experiments

We evaluate the presented approach on the MPI Sintel [6] and KITTI 2015 [24] benchmarks. When reporting experimental results, we refer to our approach as DC Flow. Feature computation, cost volume construction, and cost volume processing were implemented in OpenCL and evaluated on an Nvidia TITAN X GPU. Postprocessing is performed on an Intel Xeon E5-2699 CPU. Unless stated otherwise, a 64-dimensional feature embedding was used.

**MPI Sintel.** MPI Sintel is a challenging dataset with large displacement, motion blur, and non-rigid motion [6]. The public training set consists of 23 sequences of up to 50 images each. We randomly select 14 sequences from the *final* rendering pass for training, and use the remaining 9 sequences as a validation set.

Table 1 compares our result to prior work on the final pass of the test set. All errors are measured as average end-point error (AEPE). We use the 9 standard metrics [6], which evaluate the average EPE over different subsets of the image: all pixels, non-occluded pixels (noc), occluded pixels (occ), pixels within a given range of distances to the nearest occlusion boundary (d0-10, d10-60, d60-140), and pixels with a velocity in a given range (s0-10, s10-40, s40+). At the time of writing, our approach is ranked first on the Sintel leaderboard. We outperform all competing methods on seven out of nine evaluation metrics, including the main one (all). Our approach performs particularly well in regions that undergo fast motion (s40+). Qualitative results on the validation set are shown in Figure 3.

| Method | all | noc | occ | d0-10 | d10-60 | d60-140 | s0-10 | s10-40 | s40+ |
|---|---|---|---|---|---|---|---|---|---|
| PatchBatch [12] | 6.783 | 3.507 | 33.498 | 6.080 | 3.408 | 2.103 | **0.725** | **3.064** | 45.858 |
| EpicFlow [27] | 6.285 | 3.060 | 32.564 | 5.205 | 2.611 | 2.216 | 1.135 | 3.727 | 38.021 |
| DiscreteFlow [25] | 6.077 | 2.937 | 31.685 | 5.106 | 2.459 | 1.945 | 1.074 | 3.832 | 36.339 |
| CPM-Flow [18] | 5.960 | 2.990 | 30.177 | 5.038 | 2.419 | 2.143 | 1.155 | 3.755 | 35.136 |
| FullFlow [7] | 5.895 | 2.838 | 30.793 | 4.905 | 2.506 | 1.913 | 1.136 | 3.373 | 35.592 |
| SPM-BPv2 [21] | 5.812 | 2.754 | 30.743 | 4.736 | 2.255 | 1.933 | 1.048 | 3.468 | 35.118 |
| DDF [13] | 5.728 | 2.623 | 31.042 | 5.347 | 2.478 | 1.590 | 0.959 | 3.072 | 35.819 |
| FlowFields+ [3] | 5.707 | 2.684 | 30.356 | 4.691 | 2.117 | 1.793 | 1.131 | 3.330 | 34.167 |
| DC Flow | **5.119** | **2.283** | **28.228** | **4.665** | **2.108** | **1.440** | 1.052 | 3.434 | **29.351** |

Table 1. Comparison to state-of-the-art optical flow methods on the **Sintel *final* test set** in terms of AEPE. At the time of writing, our approach is ranked first on the Sintel leaderboard. We outperform competing methods on seven out of nine evaluation metrics, including the main one (all).

| Method | Domain-agnostic | Non-occluded pixels (%) | | | All pixels (%) | | | Runtime |
|---|---|---|---|---|---|---|---|---|
| | | Fl-bg | Fl-fg | Fl-all | Fl-bg | Fl-fg | Fl-all | |
| SOF [31] | ✗ | 8.11 | 18.16 | 9.93 | 14.63 | 22.83 | 15.99 | 6 min |
| JFS [19] | ✗ | 7.85 | 14.97 | 9.14 | 15.90 | 19.31 | 16.47 | 13 min |
| SDF [2] | ✗ | 5.75 | 18.38 | 8.04 | 8.61 | 23.01 | 11.01 | – |
| EpicFlow [27] | ✓ | 15.00 | 24.34 | 16.69 | 25.81 | 28.69 | 26.29 | 15 sec |
| FullFlow [7] | ✓ | 12.97 | 20.58 | 14.35 | 23.09 | 24.79 | 23.57 | 4 min |
| CPM-Flow [18] | ✓ | 12.77 | 18.71 | 13.85 | 22.32 | 22.81 | 22.40 | 4.2 sec |
| DiscreteFlow [25] | ✓ | 9.96 | **17.03** | 11.25 | 21.53 | **21.76** | 21.57 | 3 min |
| DDF [13] | ✓ | 10.44 | 21.32 | 12.41 | 20.36 | 25.19 | 21.17 | 1 min |
| PatchBatch [12] | ✓ | 10.06 | 22.29 | 12.28 | 19.98 | 26.50 | 21.07 | 50 sec |
| DC Flow | ✓ | **8.04** | 19.84 | **10.18** | **13.10** | 23.70 | **14.86** | 8.6 sec |

Table 2. Comparison to state-of-the-art optical flow methods on the **KITTI 2015 test set**. Our domain-agnostic approach outperforms prior such methods by a significant margin, on both occluded and non-occluded pixels. The presented approach outperforms the most accurate prior method on the main Fl-all measure by 29.5%. For completeness, we list recent domain-specific methods at the top of the table. The presented approach outperforms two of these methods without using domain-specific information.

**KITTI 2015.** KITTI 2015 is an automotive dataset of road scenes [24]. It contains 200 training images with semi-dense ground-truth flow. We withheld 30 randomly selected images for validation and trained the feature embedding on the remaining 170 images.

A comparison to prior work on the KITTI 2015 test set is provided in Table 2. Following the standard protocol on this dataset, we report the percentage of pixels with an EPE above 3 pixels. The table reports the standard measures on this dataset: error over the static background (Fl-bg), error on dynamic objects (Fl-fg), and error over all pixels (Fl-all). The three measures are reported for all pixels as well as non-occluded pixels. The primary evaluation measure is Fl-all over all pixels. Our approach yields an error of 14.86 according to this measure, which is 29.5% lower than the most accurate prior domain-agnostic method (PatchBatch). On non-occluded regions we outperform the most accurate domain-agnostic method (DiscreteFlow) by 9.5%, which indicates that our approach derives its advantage from both better matches and a better inpainting procedure. Our approach is particularly accurate in background regions and delivers competitive performance in foreground regions.

For completeness, Table 2 (top) lists the performance of recent methods that use additional domain-specific semantic information to enhance their optical flow estimates. These methods are expected to perform better than domain-agnostic approaches on this benchmark, at the cost of generality. Nevertheless, our approach outperforms two of these recent methods and is only surpassed by one domain-specific pipeline [2], without using domain-specific information. Example results on the validation set can be seen in Figure 4.

**Ablation study.** We conduct experiments on the validation sets of both Sintel and KITTI 2015 to evaluate the contribution of different components of the presented approach. For all experiments we provide results for two different settings of the effective search range: a fast version ($r_{max} = 100$) and an accurate version ($r_{max} = 242$). We report AEPE over all pixels for Sintel and percentage of wrongly matched

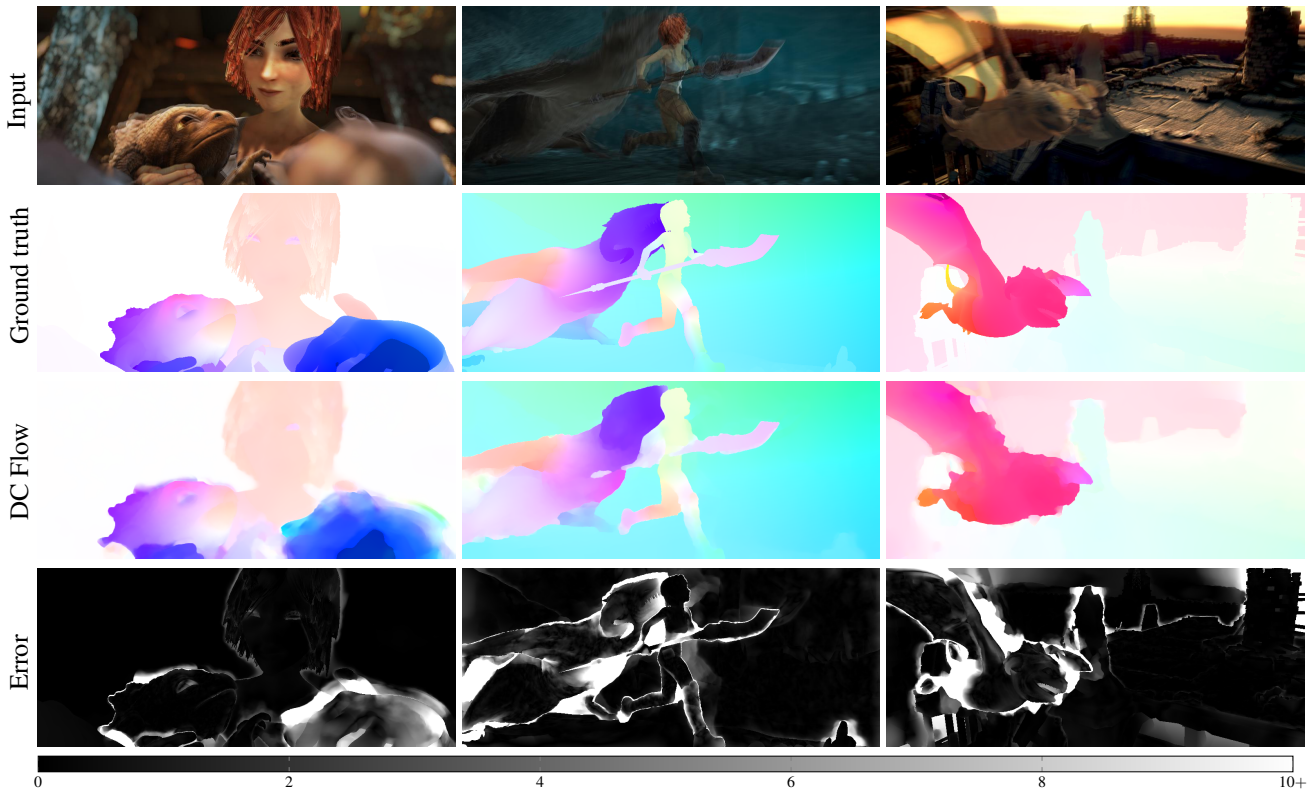

Figure 3. Qualitative results on three images from the Sintel training set. From top to bottom: superimposed input images, ground-truth flow, optical flow computed by the presented approach, corresponding EPE maps, and color code of the EPE maps.

pixels in occluded and non-occluded regions for KITTI.

We first conduct a controlled experiment to demonstrate the effectiveness of the learned feature embedding. We use a feature dimensionality of $d = 64$ and construct the cost volume as described in Section 4. To isolate the learned feature embedding from the rest of the presented pipeline, we pass the constructed cost volume to Full Flow [7]. This replaces the classical NCC matching function used in that work by our learned feature embedding, while keeping the rest of that pipeline fixed. The results are reported in Table 3 (top). Our feature embedding (Ours+FullFlow) yielded consistently lower error than the classical NCC cost (NCC+FullFlow), on both datasets.

Next, we focus on the cost volume processing and postprocessing, presented in Sections 5 and 6. The results are reported in Table 3 (bottom). The matches provided by our cost volume are sufficiently accurate for naive winner-takes-all selection with no cost volume processing (Ours+WTA) to yield respectable accuracy, approaching the complete Full Flow pipeline, which includes global optimization. (In the Ours+WTA condition, 97% of the running time is consumed by EpicFlow interpolation.) Adding Flow-SGM to our pipeline (Ours+SGM) further increases accuracy and even surpasses the corresponding Ours+FullFlow variants reported at the top of the table.

Adding homography-based inpainting in the postprocessing stage (Ours+SGM+H) maintains high accuracy on Sintel and significantly improves accuracy on KITTI. The difference in the effect of the postprocessing stage on the two benchmarks is not surprising given the mostly rigid nature of KITTI scenes, which makes them particularly amenable to homography fitting.

The influence of feature dimensionality is shown in Table 4. Surprisingly, feature embedding with dimensionality as low as 10 performs remarkably well and could be used in practice.

| Method | Sintel AEPE | KITTI 2015 noc (%) | occ (%) | Time (sec) |
|---|---|---|---|---|
| NCC+FullFlow (fast) | 6.91 | 16.09 | 25.11 | 40 |
| NCC+FullFlow (acct) | 6.37 | 14.33 | 23.48 | 240 |
| Ours+FullFlow (fast) | 6.31 | 12.74 | 22.17 | 20 |
| Ours+FullFlow (acct) | 6.01 | 11.10 | 20.40 | 120 |
| Ours+WTA | 7.22 | 18.06 | 27.37 | 3.0 |
| Ours+SGM (fast) | 6.08 | 12.78 | 22.46 | 3.4 |
| Ours+SGM (acct) | 5.51 | 10.72 | 20.47 | 5.7 |
| Ours+SGM+H (acct) | **5.44** | **10.21** | **15.09** | 8.6 |

Table 3. Controlled experiments that evaluate the contribution of different components of the presented approach. Top: evaluation of the learned feature embedding. Bottom: the effect of Flow-SGM and homography-based inpainting.

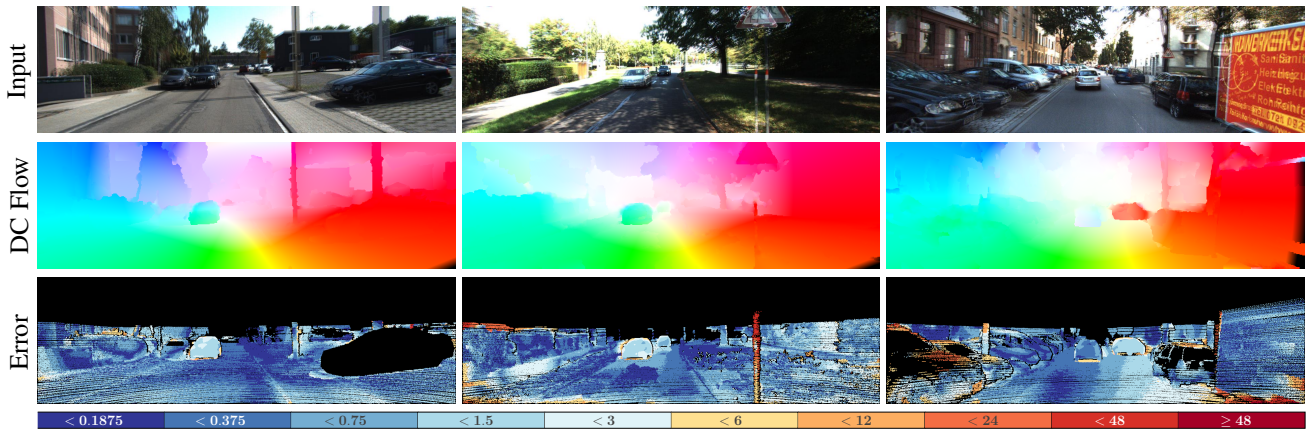

Figure 4. Qualitative results on three images from the KITTI 2015 training set. From top to bottom: superimposed input images, optical flow computed by the presented approach, corresponding error maps, and color code for error maps. Colors indicate error thresholds.

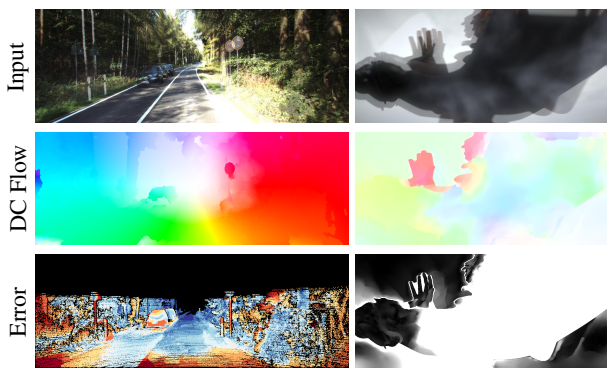

Figure 5. Failure cases. An example from the KITTI 2015 dataset on the left, an example from Sintel on the right.

A breakdown of the running time for each component of the presented approach is shown in Table 5. Cost volume construction is nearly real-time (80 milliseconds for both directions) in the 'fast' condition and is still extremely rapid (260 milliseconds) in the 'accurate' condition. In the 'fast' condition, the running time is dominated by EpicFlow inpainting (84% of the runtime). (Homography inpainting is not used in this condition.) In the 'accurate' condition, cost volume processing takes roughly $\frac{1}{3}$ of the total running time and postprocessing consumes the other $\frac{2}{3}$.

| Dimension | Sintel AEPE | KITTI 2015 noc (%) | occ (%) |
|---|---|---|---|
| 10 | 5.71 | 11.70 | 21.42 |
| 16 | 5.64 | 11.43 | 21.29 |
| 32 | 5.53 | 11.10 | 20.75 |
| 64 | **5.51** | **10.72** | **20.47** |

Table 4. Effect of the feature dimensionality on accuracy.

|  | fast | accurate |
|---|---|---|
| Feature extraction | 0.02 | 0.02 |
| Cost volume (fwd + bwd) | 0.06 | 0.24 |
| SGM (fwd + bwd) | 0.45 | 2.59 |
| EpicFlow | 2.87 | 2.87 |
| Homography inpainting | – | 2.91 |
| Total | 3.40 | 8.63 |

Table 5. Running time for each component of the presented approach (seconds).

Finally, some failure cases are shown in Figure 5. On Sintel, failure cases are typically due to dramatic occlusion, strong motion blur or large motion of untextured objects. On KITTI, most failure cases are due to shading and overexposed regions.

## 8. Conclusion

We have presented an optical flow estimation approach that directly constructs and processes the four-dimensional cost volume. We have shown that, contrary to widespread belief, a highly accurate cost volume can be constructed in a fraction of a second. To this end, we use a learned feature embedding. The constructed cost volume is processed using an efficient adaptation of semi-global matching to the four-dimensional setting. Our approach is rooted in classical stereo estimation approaches that have been widely deployed and thoroughly tested in the field. Our work makes a step towards unifying optical flow and stereo estimation, which have hitherto been separated by computational considerations despite the structural similarity of the problems. Our approach combines high accuracy with competitive runtimes, outperforming prior methods on standard benchmarks by significant margins.